\documentclass[utf8]{hidden}

\usepackage{url,hyperref,lineno,microtype,subcaption}
\usepackage[onehalfspacing]{setspace}
\usepackage{amsmath}

\usepackage{bm}
\usepackage{hyperref}
\usepackage{draftwatermark}
\usepackage{natbib}

\usepackage{enumitem}
\usepackage{tabularx}

\def\keyFont{\fontsize{8}{11}\helveticabold }
\def\firstAuthorLast{Simmons {et~al.}} 
\def\Authors{Joe Simmons\,$^{1}$,Paul Bremner\,$^{2,*}$, Thomas J Mitchell\,$^{3,4}$, Alison Bown\,$^{1}$ and Verity M\textsuperscript{c}Intosh\,$^{1,4}$}

\def\Address{$^{1}$Bristol VR Lab, University of the West of England, Bristol, UK \\
$^{2}$Bristol Robotics Laboratory, University of the West of England, Bristol, UK \\
$^{3}$Digital Cultures Research Centre, University of the West of England, Bristol, UK\\
$^{4}$Creative Technologies Lab, University of the West of England, Bristol, UK}

\def\corrAuthor{Corresponding Author}

\def\corrEmail{paul2.bremner@uwe.ac.uk}

\begin{document}
\onecolumn
\firstpage{1}

\title[The Ballad of the Bots]{The Ballad of the Bots: Sonification Using Cognitive Metaphor to Support Immersed Teleoperation of Robot Teams} 

\author[\firstAuthorLast ]{\Authors} 
\address{} 
\correspondance{} 

\extraAuth{}

\maketitle

\begin{abstract}

\section{}
As an embodied and spatial medium, virtual reality is proving an attractive proposition for robot teleoperation in hazardous environments. This paper examines a nuclear decommissioning scenario in which a simulated team of semi-autonomous robots are used to characterise a chamber within a virtual nuclear facility. This study examines the potential utility and impact of sonification as a means of communicating salient operator data in such an environment. However, the question of what sound should be used and how it can be applied in different applications is far from resolved. This paper explores and compares two sonification design approaches. The first is inspired by the theory of cognitive metaphor to create sonifications that align with socially acquired contextual and ecological understanding of the application domain. The second adopts a computationalist approach using auditory mappings that are commonplace in the literature. The results suggest that the computationalist approach outperforms the cognitive metaphor approach in terms of predictability and mental workload. However, qualitative data analysis demonstrates that the cognitive metaphor approach resulted in sounds that were more intuitive, and were better implemented for spatialisation of data sources and data legibility when there was more than one sound source.

\tiny
 \keyFont{ \section{Keywords:} keyword, keyword, keyword, keyword, keyword, keyword, keyword, keyword} 
\end{abstract}

\section{INTRODUCTION}
In an era of rapid technological convergence and advancement, many industries are investing in new strategies to reform high risk practices, and to reconsider the management of hazardous tasks in industrial settings. Settings in which direct human intervention might be considered unsafe or inappropriate. For example, in applications ranging from underwater surveillance to disaster relief, tools such as immersive technologies, artificial intelligence and autonomous robot systems are starting to be combined to give remote operators the sense of being `close to the action' whilst engaging at a safe physical distance. 

In many cases virtual reality (VR) may be considered a desirable tool due to the psychological sensation of \textit{presence}, which here we define as a sense of ``being there'' within remote environments (as defined in \citep{murphy2020we}). A sense of presence can elevate levels of trust \citep{Salanitri2016}, which can be vital in the operation of semi-autonomous robots where operators need to rely on correct behaviours. Perhaps more importantly the affordances associated with presence of embodiment and kinaesthesia have been shown to support users of teleoperation systems in task performance and decision making, seemingly through enhanced spatial awareness and sensory cognition in comparison with more traditional teleoperation interfaces such as 2D video and text \citep{whitney2019comparing}. 

In this study researchers worked in partnership with Sellafield Ltd to explore a specific nuclear decommissioning scenario in which virtual reality (VR) is used to teleoperate a simulated team of semi-autonomous robots as they characterise and map a chamber with largely unknown contents within a virtual nuclear facility. 
In this scenario the operator is responsible for interpreting sensor data gathered by the robots, identifying invisible hazards so they can be labelled. Operators are also responsible for monitoring and protecting the robots from the negative effects of prolonged exposure to such hazards. The study scenario was co-designed with Sellafield workers to represent a plausible use case for near future working in which semi-autonomous robot teams are guided by a remote user to characterise potentially hazardous spaces using a VR digital twin environment built from robot sensor data (as opposed to the entirely simulated environment used here). 

This study specifically examines the potential utility and impact of sound and sonification in the design of such systems. The human auditory system has a high temporal resolution, wide bandwidth and is able to localise and isolate concurrent audio streams within an audio scene \citep{carlile2011}. Our approach leverages best practice from immersive systems design, using the 360 degree nature of sonic experience to reduce the ``sensory overload'' caused by an over-reliance on visual systems \citep{Benyon2005}, and to enable a user to engage in ``attentional switching to best achieve aims and monitor for potentially salient distractions'' \citep{kurta2022targeting}. Building on preliminary studies \citep{Bremner2022,Simmons2023}, we leverage the capacity of VR as an embodied, multi-modal tool, spreading salient live data across visual and auditory channels, giving operators much of the information that they need to complete tasks in the form of carefully designed bespoke sonification sound sets without the need for overloading the visual channel with this information. Further, sound has previously been shown to enhance users' sense of presence in virtual environments \citep{larsson2007you}, a desired condition as outlined above.  

The bespoke sound sets used in this study have been created by professional sound designers to exemplify two distinct design approaches to sonification. The first sound set applies the principles of cognitive metaphor, engaging with theories of embodied image schema theory, narrative scaffolding and cognitive metaphor theory \citep{Roddy2020, Wirfs2021}. This sound set has been developed in response to the literature (detailed in section \ref{sec:Background}) suggesting that naturalistic mappings may be experienced more intuitively by users, allowing them to apply socially acquired contextual knowledge that could make sounds automatically decodable and facilitate meaning making \citep{Kantan2021, Wirfs2021}. 

The second sound set follows a more traditional approach to sonic information design, assigning auditory parameters such as pitch, loudness and signal brightness to task-critical features in the environment on the assumption that their sonic distinctiveness will create a legible and functional sonic ecosystem \citep{Dubus2013}. 

Within this emergent industrial use case, and contrasting two distinct sonification design approaches, our work has significant novel contributions in the following areas of complementary interest:

\begin{itemize}
\item The capacity of a user of a VR teleoperation system to interpret and apply task data conveyed by sonification.
\item The application of cognitive metaphor approaches (from sound design philosophy) to data sonification.
\item The effect of the two sonification design strategies on task performance and user experience.  
\item Levels of trust, comfort, stress and habituation experienced by participants using a system with data sonification. 
\end{itemize}

Our work demonstrates the utility of data sonification in VR teleoperation systems, and provides important insights into the efficacy of two different sonification design approaches.

In this paper we first present the background of what sonification is, current design approaches, as well as review related literature of sonification applications in VR and HRI. We then detail our hypothesis and the scenario used. Following this we describe the VR user interface for teleoperation of simulated robots in a virtual environment, consisting of visual and audio components; within which we describe the two compared sonification design approaches. The two sonification design approaches are evaluated in two user studies: the first evaluates user experience of them within the VR environment, the second is an online study of the objective evaluation of the efficacy of our sonifications in hazard identification. Finally we discuss our results and provide conclusions of our findings. 

\section{Background}
\label{sec:Background}
\subsection{Data Sonification}

Sonification is broadly understood as ``the use of non-speech audio to convey information'' \citep{Kramer2010}, and has been used effectively in a wide range of applications: sensory substitution \citep{Meijer1992}, accessibility \citep{holloway2022}, medical diagnosis \citep{walker2019}, science communication \citep{sawe2020}, sound synthesis \citep{mitchell2020} and data-inspired music \citep{mitchell2019}. However, the question of what form of `non-speech audio' should be used and how it can be applied in different applications is far from resolved. Roddy and Bridges \citep{Roddy2020} introduce ``the mapping problem, a foundational issue which arises when designing a sonification, as it applies to sonic information design'' and refer to the aesthetic and structural challenges of parameter mapping sonification as a means to legibly communicate information. Dubus and Bresin's meta-analysis highlighted that dominant forms of parameter mapping sonification have tended towards computational approaches by assigning discrete data sources to sonic dimensions. The dimensions tend to be those common to Western tonal music such as loudness, pitch, timbre, and tempo \citep{Dubus2013, Grond2011}. Flowers argues that the simultaneous plotting of numerous continuous variables, particularly using arbitrary sonic characteristics such as pitch, is unlikely to be meaningful to general listeners \citep{Flowers2005}. 

It is important to extend sonification design research beyond how data is expressed as sound and consider how auditory displays are interpreted by listeners, considering factors relating to human perception and aesthetics. There is some debate on the role of aesthetics in sonification, with several authors stating that an auditory display becomes decreasingly useful as it becomes increasingly pleasant and musical \citep{gresham2012}, \citep{ferguson2019}. Vickers brought this dichotomy into question \citep{vickers2016}, and, along with others \citep{grond2014}, \citep{kramer1994} argues that aesthetic design strategies have the potential to reduce fatigue while enhancing the usability and expressive qualities of a display. This observation has led to numerous calls for interdisciplinary collaboration to encourage the integration of artistry and craft into sonification research \citep{barra2002}, \citep{Barrass2012}. 

Wirfs-Brock et al \citep{Wirfs2021} note that one of the ``reasons sonification has failed to reach broad audiences are the tension between cognitive versus ecological approaches to sound design''. Roddy and Bridges \citep{Roddy2020} suggest that ``the mapping problem can be addressed by adopting models of sound which draw from contemporary theories of embodied cognition to refine the more traditional perspectives of psychoacoustics and formalist/computationalist models of cognition''. In this paper we explore a model of sound which is based on the embodied cognition notion of cognitive metaphor and compare the approach with a more typical computationalist sonification approach.

\subsection{Sonification Design and Cognitive metaphor}
\label{sec:CogMet}
Sound choice in parameter mapping sonification is of key importance for conveying information, the choice of sound should be meaningful and relate to the underlying data of a parameter mapping \citep{ferguson2018investigating} and help to reduce listener fatigue and annoyance \citep{vickers2014sonification}. The use of cognitive metaphor (or conceptual metaphor \citep{lakoff2006conceptual}) draws from the field of embodied cognition, a counter-Cartesian philosophical position that holds that ``cognitive processes are deeply rooted in the body’s interactions with the world'' and that sense-making is motivated by multi-sensory, embodied experience \citep{Wilson2002}.

In sonification, embodied cognition principles can be applied in the selection of more naturalistic or contextually linked mappings that are built on the principles of conceptual metaphors \citep{Kantan2021}. Such mappings take into account the embodied associations that humans learn over a lifetime of experience \citep{Kantan2021}, although Roddy and Bridges \citep{Roddy2020} caution that these embodied associations (or embodied knowlege) can vary between different groups and cultures, recommending that “[s]uch factors must be taken into account during phases of design'' and recommending a user-centred design approach in order to create systems that are tailored to the expected user groups. Indeed, closer associations between sounds and the data that they represent could allow users to reduce cognitive load and memory burdens associated with sonification interpretation \citep{preziosi2017remembering}.

Wirfs-Brock et al \citep{Wirfs2021} suggest that a core part of the sonification design process should be ``to teach audiences to listen to data'', and that alongside cognitive metaphor, designers should consider what ``narrative scaffolding'' might be needed to support listeners as they learn to work with the complexities of a new sonic landscape. Roddy and Bridges \citep{Roddy2020} propose the Embodied Sonification Listening Model (ESLM) as a means to track and understand not just the mapping of parameters to data sources, but the mapping process that listeners undertake when arriving at their own understanding of the information received. The model builds on ideas of the `circuit of communication’ in Reception Theory in which the producer encodes their intentions within a media product, and invites the viewer to `decode' the work at distance, drawing on their own cultural frameworks and embodied knowledge to unlock layers of meaning \citep{Hall1980}. 

Further to the use of cognitive metaphor, certain design strategies vaunted for interactive technologies and systems design have proved highly relevant to this research. 

From the literature, and drawing on experience from early prototyping, we became aware that sonifications in this context could rapidly become congested and cacophonous, as multiple sound sources compete for attention. Similar to Hermann et al. \citep{Hermann2011}, Case \citep{Case2016} reminds us of ``the limited bandwidth of our attention'', recommending five core principles for the design of calm technology:

\begin{itemize}
    \item Technology should require the smallest possible amount of attention.
    \item Technology should inform and create calm.
    \item Technology should make use of the periphery.
    \item Technology should amplify the best of technology and the best of humanity.
    \item Technology can communicate, but doesn't need to speak.
\end{itemize}  

Case calls for designers to adopt aesthetics that make sense in the environment in which they are encountered, and that call attention to themselves only when relevant and of use to the person who is expected to attend to that call.

Finally, in `Designing Interactive Systems', Benyon \citep{Benyon2005} discusses the importance of mapping the information architecture within interactive systems, and paying attention to the relationship between physical and conceptual objects when designing in 3D space. Simply put “a good mapping between conceptual and physical objects generally results in better interaction.”\citep{Benyon2005}

\subsection{Sonification In Human Robot Interaction and Virtual Reality}
In the context of robotics, Zahray et al. \citep{Zahray2020} and Robinson et al. \citep{Robinson2021} have used sonification to supplement visual information on the motion of a robot arm. Both studies showed that different sounds can alter the perception of the motion as well as the robot capabilities.

Hermann et al. use sound to convey information in a complex process monitoring system of a cognitive robot architecture \citep{Hermann2003}. By dividing data features between the visual and auditory channels, operators were shown to establish a better understanding of system operation. This is conceptually similar to the aims of the work presented here.

Triantafyllidis et al. evaluate the performance benefits of stereoscopic vision, as well as haptic and audio data feedback on a robot teleoperation pick and place task; both haptic and audio feedback were collision alerts rather than for data representation as evaluated here \citep{triantafyllidis2020study}. They found that stereoscopic vision had the largest benefit, with audio and haptics having a small impact. Similarly, Lokki et al. demonstrate that users can use sonified data to navigate a virtual environment \citep{Lokki2005}. By comparing audio, visual, and audiovisual presentation of cues, they found that audiovisual cues perform best, demonstrating multimodal cue integration. 

It is clear that data sonification is an under explored area in HRI and telerobotics, with relatively few studies across a small number of application domains. Hence, the work we present here represents an important early step in understanding the utility of sonification in HRI. We are interested in understanding the utility of sonification in this novel problem domain, consequently we collaboratively design and compare two different sets of sonifications and systematically investigate their impact on task performance and workload of teleoperators. The two sets of sounds are cognitive metaphor based, i.e more literal, calling on established representations of these parameters, and computationalist based, with no direct connection to established sound representations. 

\section{Methodology}

\subsection{Hypotheses}

Drawing on the literature we will explore the impact of using a cognitive metaphor approach to sound design (cog) vs a more traditional computationalist approach (comp), and hypothesise that sonifications designed in line with principles of cognitive metaphor will: 
\begin{enumerate}[itemindent=0.5cm, labelsep= 0.3cm, label=\textbf{H\arabic*}]
    \item increase presence/immersion
    \item increase trust
    \item be easier to interpret and reduce workload
    \item reduce listener annoyance and fatigue
    \item meaningfully relate to the underlying data types
    \item will make data level understanding more difficult
\end{enumerate}

To test these hypotheses a between-subjects study was conducted in order to compare the effect of two sonification approaches that are intended to convey real-time measurements of radiation, temperature and gas. The first `cognitive metaphor' approach draws, where possible, from embodied and established associations that humans already have with the underlying data types. The second `computationalist' approach disregards any established experience and adopts the most prevalent parameter mappings found in the literature. We anticipate that by using a sonification design approach that is ecologically linked with the environment the sense of presence \citep{larsson2007you,kern2020audio} and consequently trust will be increased \citep{Salanitri2016} (H1 \& H2). Sounds that more closely align with user expectations will be easier to interpret and will reduce the cognitive load \citep{schewe2020ecological} (H3). Sonification mappings that take an ecological design approach inspired by cognitive metaphor will improve the user experience by reducing listener annoyance and fatigue when compared with mappings to common auditory parameters, which are designed to efficiently convey arbitrary information (H4). Aligning the sonification mappings with user expectations will make the data types will be easier to identify \citep{walker2005mappings, ferguson2018investigating} (H5). We expect the data levels to be easier to understand with the utilitarian approach since there is a clearer mapping between sound and data with no consideration taken for listener comfort or environmental coherence (H6). However, our aim is to design both sound sets with understanding in mind, meaning that this difference should be minimal.

\subsection{The Scenario}

A real world industrial use case developed with our partners at Sellafield Ltd. is that of characterisation and mapping of sites to be decommissioned. There exist many legacy sites for which the exact physical and environmental conditions are unknown, necessitating this characterisation process. An approach proposed to solving this challenge developed on the Robots for Nuclear Environments programme grant is the use of multi-robot teams \citep{web:RNE}. In order to comply with the nuclear safety case human oversight is required in this process to ensure the environment is safely and correctly mapped (autonomous behaviours may have to be overridden to gather needed data), and decommissioning activities properly planned. Hence, the robots need to be teleoperated, and their gathered data interpreted by a human operator. The approach being developed is to use semi-autonomous robots to gather environment data in order to construct a digital-twin of the environment which can be safely navigated in VR, and used to allow the operator intuitive understanding of the environment and to issue commands to the robots. Within the environment there are hazardous conditions of which the operator needs to be aware in order to safely and usefully direct the robots and plan decommissioning activities. We have adapted this scenario to allow us to evaluate sonification approaches by sonifying hazard data so that an operator can be aware of hazard levels while still being able to visually focus on the physical environment and the activities of the robots.

In order to allow us to evaluate the sonification approaches while the robot teleoperation and environment reconstruction systems are being developed we have designed a simulated nuclear environment within which a characterisation and mapping task can be conducted. It simulates object reconstruction based on 3D scans of real objects, and sensor readings of environmental hazards (which are sonified). A simulated (and simplified) robot controller and teleoperation interface is used to direct the robots when needed. Thus, we have a facsimile of a digital twin even though no real environment or robots exist, we term this a simulated twin. 

\begin{figure*}
    \centering
    \includegraphics[width=\textwidth]{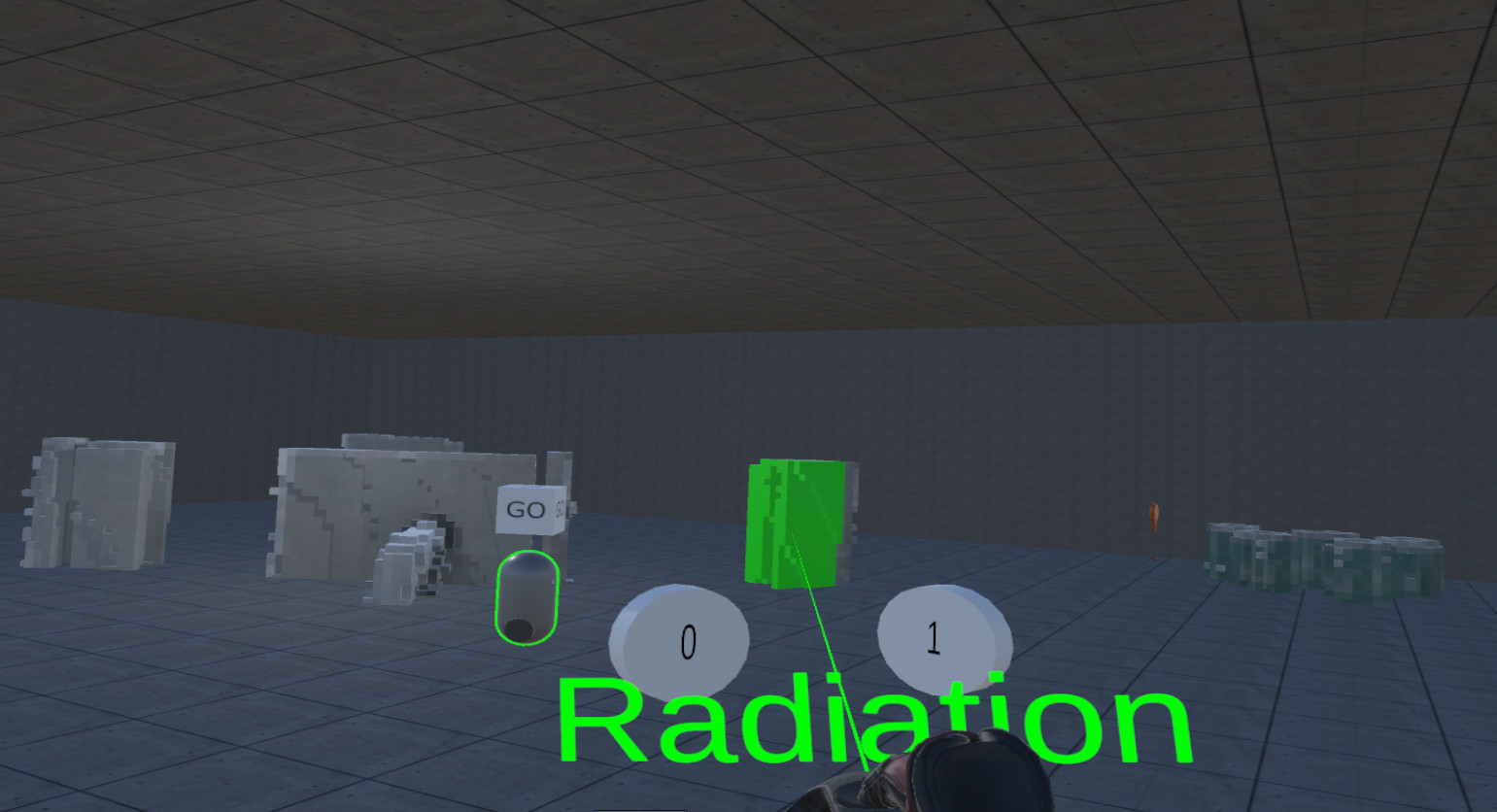}
    \caption{The virtual reality environment. Voxel objects that have been revealed by the robots can be seen and one is tagged as radioactive. The laser pointer is in radiation tagging mode. A robot is selected in waypoint mode (green outlined robot), and two waypoints have been set. A hazard marker is visible by the barrels on the right.}
    \label{fig:GUI}
\end{figure*}

\section{Multi-modal Interaction Design for Robot Teleoperation}

We have developed a virtual reality 'simulated twin' environment for our use case. The virtual environment enables  data observation and robot teleoperation within a simulated nuclear decommissioning (ND) scenario. In this scenario, a team of four robots move autonomously to 'scan' and map an unknown chamber of a nuclear facility. The robots have virtual 'sensors' that detect radiation, temperature, and flammable gas, environmental hazards detailed as important by experts from Sellafield Ltd. The virtual environment is comprised of a graphical user interface (GUI) for observing robot sensor data and operating the robots, and a data sonification system for relaying as sound data from simulated hazard sensors. Users must use the robot's sensor data to characterise areas of high hazard levels (tag objects with an appropriate data label) for subsequent stages of the ND process. An additional goal for users of the system is to keep the robots as safe as possible to minimise damage to them, hence reducing the frequency of repair and replacement costs; further, inoperative robots that have become irradiated become additional waste that must be disposed of during decommissioning. In the following sections we outline the simulated environment, the user interface needed for the tasks, and the audio implementation. A more detailed description of the task is provided in section~\ref{sec:StudyDesign}. A video of system operation can be found \href{https://youtu.be/vPuy7autXCc}{\color{blue}\underline{here}}.

\subsection{Graphical User Interface}
The graphical user interface (GUI) was built using Unity, utilising the SteamVR Unity package for additional assets and features chosen to resemble Sellafield nuclear facility. The GUI includes a simulated ND environment which is explored by a team of robots along with task performance interfaces needed for interacting with the environment and robots. An overview of the GUI is shown in Figure~\ref{fig:GUI}.

\subsubsection{Environment Design}
The environment comprises voxel objects that are initially invisible, revealed when detected by the proximity sensors of passing robot(s). The objects have been produced by scanning real-world simulacrums of nuclear decommissioning objects using an RGB-D camera, the resulting pointclouds are then converted to voxel renderings. The floor and walls are always visible, representing information known \textit{a priori}. This setup simulates a typical nuclear decommissioning scenario where the floor plan of the rooms is available from blueprints but the exact location of objects and hazards is unknown.

\begin{figure}
    \centering
    \includegraphics[width=\columnwidth]{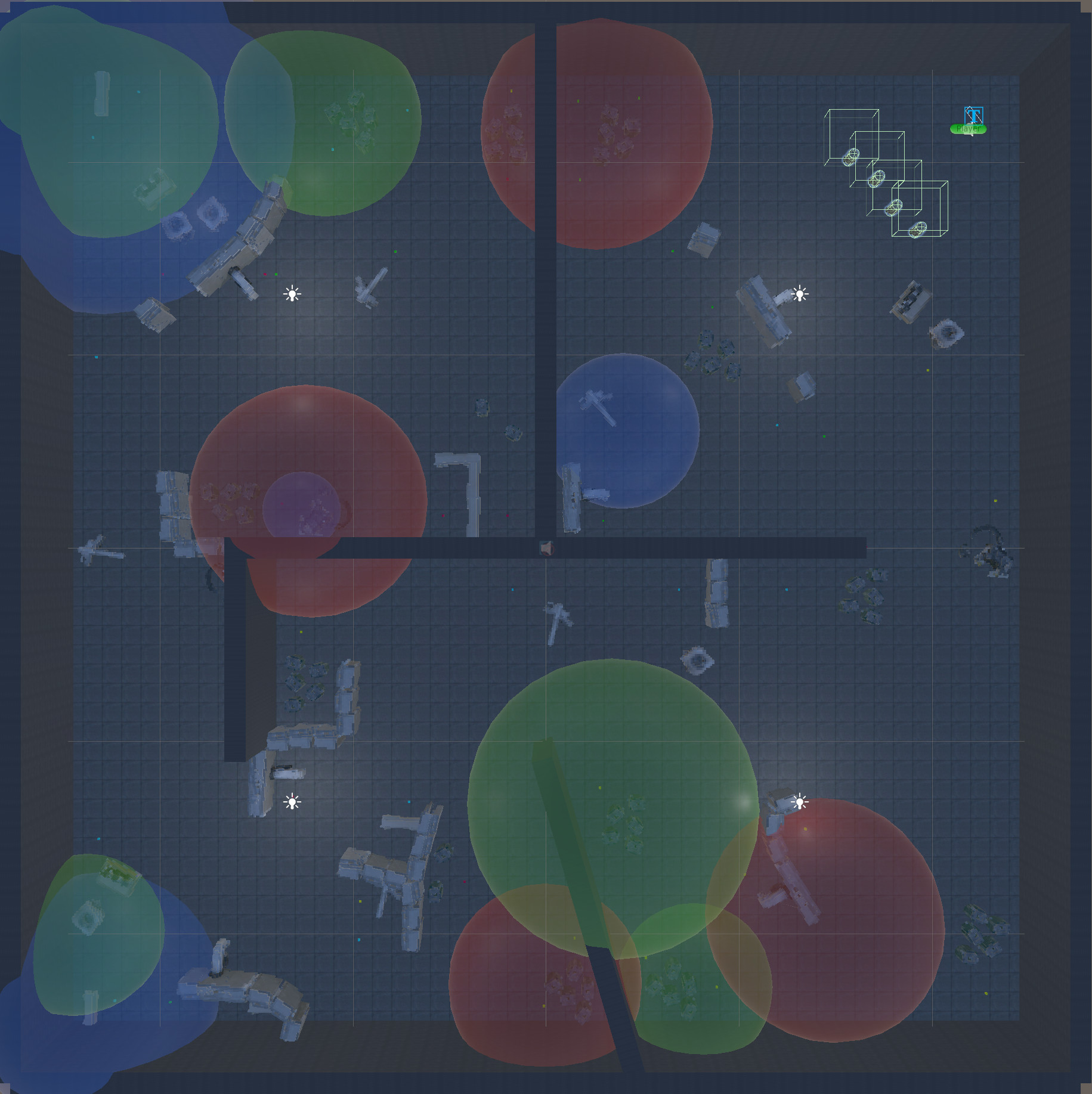}
    \caption{Layout of the simulated nuclear environment. Hazard spheres are made visible and are red for temperature, green for radiation, and blue for flammable gas. The robots and the user avatar start in the top right-hand corner.}
    \label{fig:map}
\end{figure}

In order to simulate hazards in the VR environment that are not solid objects, the space is populated with invisible hazard spheres within which hazard levels increase linearly from the edge to the centre. Hazard spheres exist for each of the three types of hazard, with a variety of sizes and positioning, including overlapping spheres (which for matching hazard types are additive up to a maximum level), and are deployed to create hazardous areas in the environment for the user tasks. The hazard level at a given point in space drives the output of the sonification system described in section~\ref{sec:audio}. A birds-eye view of the environment with hazard spheres made visible is shown in Figure~\ref{fig:map}

\subsubsection{User Navigation}
Following the conventions of contemporary, virtual reality experience design, users have six degrees of freedom (6DOF), and can freely walk in their immediate environment with a 1 to 1 mapping between their physical and virtual location. To support navigation over longer distances, two further movement modes (both commonly deployed in VR experiences) are also made available. The first is the teleportation system provided by SteamVR: the user holds a button on one of the VR controllers and this projects an arc with a target point where the arc collides with the floor, when the target is placed as desired the user releases the button and teleports to that location. The teleportation arc is blocked by visible objects, i.e., when an area is unexplored by robots it appears to contain no objects so it can be teleported into freely. The second movement system utilises the D-pad on the VR controller. Clicking forward/backward jumps the user a short distance in that direction, allowing movement within the space without inducing nausea (as continuous controller based motion can), as well as navigation through obstacles to more easily reach specific locations. Clicking left/right jump rotates the user 45$^\circ$ in that direction, allowing the user to make large rotations without having to turn their head/body large amounts, increasing comfort and avoiding cable tangling issues.

\subsubsection{Robots - autonomous behaviour}
\label{sec:robotnavigation}
The robots use the Unity Navmesh navigation system to travel between a series of checkpoints which determine their exploration route through the environment. The Unity Navmesh system allows the navigable area to be specified, and Unity is then able to calculate the path between a Navmesh agent's current location and a specified point on the Navmesh. Initially the environment appears to be empty to the robots, so they navigate along the shortest path. As objects are revealed by the robots' sensors the Navmesh is updated to include non-traversable regions occupied by the revealed objects, and the path to the next checkpoint is updated. 

In earlier testing \citep{Bremner2022} we learned that users find it frustrating when robots appear to work against their own interests, dwelling in areas that have already been identified as hazardous at a risk to their own health. To allow the robots to avoid hazardous areas of the environment the hazard spheres are populated with Navmesh modifier tiles that can be used to increase the cost to travel through hazardous areas. When a robot detects a tile it occupies is hazardous, the Navmesh is updated to increase the cost to travel through that tile (proportionally by hazard level), and trigger path replanning. As the Unity Navmesh system uses cost to travel in its path finding, the robots will tend to avoid returning to hazardous areas unless directed to do so by the user.

\subsubsection{Robots - Control Systems}
Users are invited to engage in two main tasks: 1) utilise the robots' sensors to identify and tag hazardous areas in the environment, and 2) ensure the safety of the robots (section~\ref{sec:StudyDesign}). To support these tasks, users can issue instructions to the robots, overriding their autonomous behaviours. 

\subsubsection{Setting Waypoints}
Waypoint control utilises a laser pointer (attached to the controller in the users right hand) to place waypoints on the ground which the selected robot navigates between. Waypoints are placed by pointing and clicking at locations on the ground, the waypoints spawned at the targeted locations are numbered to indicate the order of navigation. Waypoints not yet visited can be removed with a laser pointer click. On entering waypoint control mode the robot stops moving and a 'Go' button is visible directly over the robot, navigation to waypoints may be initiated by clicking of the 'Go' button. As the robot arrives at each waypoint it is removed, waypoints can be removed by the user or by an object being discovered at the waypoints location, and this results in the robot immediately re-planning its path. When a robot is deselected it will travel to its remaining waypoints and then resume its autonomous navigation.

\subsubsection{Status indicators}
Each robot has an outline that can be made visible and is coloured to indicate a particular status; to increase utility the outline visibility is not blocked by intervening obstacles. A red outline is used to give a visual indication of a priority alert (indicating the robot is in danger, see Section~\ref{sec:priorityalerts}), an orange outline indicates real-time listening (RTL) mode is engaged (where hazard data is sonified for that robot alone, see Section~\ref{sec:RTL}), a green outline indicates the robot is in waypoint control mode (described above, RTL is also active in this mode). Real-time listening and waypoint control modes are selected for a particular robot using the laser pointer, clicking once on the robot enables RTL mode (it continues to navigate autonomously), clicking the robot again enables waypoint control mode, clicking the robot again deselects it. Only one robot may be in RTL mode at a time (to reduce audio clutter), so clicking a second robot deselects the previous robot (if one is selected).

\subsubsection{Hazardous Areas of the Environment}
\label{sec:hazards}
To support the tagging of objects that are hazardous (see section~\ref{sec:StudyDesign}), two UI elements were designed: hazard markers, and object tagging. As robots move through the environment, they detect areas of high hazard levels and on doing so place a visible marker (provided there are no other markers within a threshold radius) allowing the operator to identify areas that require further investigation. The user can use the right controller to project a laser and tag objects as hazardous (i.e., allocating hazard types from those advised by Sellafield). The laser is coloured according to the hazard tag that is to be assigned to an object: red for heat, green for radiation, blue for flammable gas, black for removing all tags. The tag colour may be toggled using the side button on the controller, and a text label appears over the controller to remind the user of the meaning of each laser colour. When the laser is pointed at a voxel chunk and a button is pressed the chunk changes colour to that of the laser. If an object is already tagged, and a new tag colour is assigned, the colours are added together using a light mixing paradigm, e.g., red and green tags make an object yellow; if the black laser is selected all tags are removed when an object is tagged. 

\subsection{Audio Implementation}
\label{sec:audio}
Whilst attending to the tasks, users experience \textit{either} the sounds designed according to principles of cognitive metaphor (cog) \textit{or} those designed in line with a computationalist approach (comp).
All audio has been implemented using Audiokinetic's Wwise and integrated into the Unity environment using C\# scripts. Unless stated otherwise, all sounds in our audio scene are spatialised, i.e., they are processed as if they emit from specified point sources within the environment. Artificial reverberation is also used to simulate the characteristics of a real environment. These global audio features not only enhance the realism of our ``cyber-physical model'' \citep{walker2022virtual}, but aid users in locating robots and navigating the virtual environment. The following audio elements are present: Real-Time Listening, Self Real-Time Listening, Notifications, Priority Alerts, UI Feedback and Ambience, each described below.

\subsubsection{Real-Time Listening}
\label{sec:RTL}
The principal sonifications in our virtual environment comprise three parameter mappings communicating the radiation, temperature and flammable gas levels measured by each robot. This sonification is the only modality by which real-time hazard levels are conveyed (by mapping hazard levels to sound parameters) and referred to as `real-time listening' (RTL). RTL is the only information users receive to perform their task of labelling highly hazardous objects. For this reason, primary focus was given to the design of these sounds, with further sonification elements designed around them.

Upon trialling RTL within a VR test environment, and hearing concurrent spatialised sonification streams for a team of up to four robots, it was clear that the auditory scene could quickly become congested and stressful for users. Degradation in the perceptual clarity of the streams was especially pronounced when multiple robots were simultaneously sensing the same kind of data type, as the user would hear multiple audio streams of the same mapping from multiple point sources. On this basis, we chose to limit live hazard data sonification (RTL) to one point source at a time by requiring the user to select a single robot on which to activate RTL mode. Once selected, the user is then able to listen to the live, sonified sensor data as though the sound is output by the selected robot. 

\subsubsection{Self Real-Time Listening}
If the user prefers to listen in to where they themselves are in the space, they may select themselves, and RTL is applied to their own avatar (i.e., they are the sound source), a practice we refer to as 'Self-RTL'. When Self-RTL is enabled, any robot RTL is deactivated to ensure only a single RTL sound-source is active.
To select this mode, the laser pointer is used to select the left hand of the user, which then acquires an orange outline to confirm this mode is active. 

In order to preserve the integrity of the defined industrial use case (where information in the simulated space is generated only by the sensors of physical robots in the nuclear facility) when the user activates Self-RTL mode, they can only hear sounds in locations that have already been mapped by robots. As the user's avatar has no physical analogue in the nuclear facility, they would be dependent on the digitally twinned robots to provide live, sonified data regarding hazard levels. To make this legible for users, when Self-RTL mode is enabled, the paths that robots have travelled through the environment are visualised. 
The navmesh modifier tiles within the hazard spheres (described in Section \ref{sec:robotnavigation}) only allow the sonifications to be audible if a robot has travelled through that tile. 

\subsubsection{Notifications}
When neither robots nor a users' own avatar are selected, RTL is turned off. However, in order that users have a continued awareness of the state of the robots (even when they are not currently visible) we designed a set of notification sounds. Users are notified when a robot first encounters a hazard, a short notification sound, or ``earcon'' \citep{dubus2013systematic}, - short snippets of RTL counterparts are emitted when they encounter a hazard. These spatialised sounds are designed to communicate both hazard type and the respective robot's relative location to the user, even if visibility is obscured. Case and Day list scenarios when it is appropriate to include notifications: ``the message is short and simple'', ``information is continually changing'', ``the user's eyes are focused elsewhere'', ``the environment limits visibility'' \citep{case2018designing}. Our notifications are designed in alignment with these criteria.

\subsubsection{Priority Alerts}
\label{sec:priorityalerts}
In order to allow users to perceive circumstances in which a robot has encountered environmental features that are hazardous to its health, we designed a set of priority alerts 
Priority levels are useful as a safeguarding measure and combine a robot's internal ``health'' and external risk factors, informing users when interventions are required to preserve robot safety. Each robot’s priority level calculations are normalised to a range of 0 to 1, with values above 0.9 triggering a high priority alert. High priority alerts are designed to take precedence over the rest of the auditory scene, but rather than making them louder, we dynamically applied gain reductions and filtering to all other audio streams to limit alarm fatigue. Even still, with three hazard priority sonifications and four robots, the user could hear up to twelve concurrent alerts, so all alerts are synchronised when simultaneously active, ensuring as much information be sonically communicated as possible without becoming cacophonous. When any robot’s priority level exceeds 0.95, an additional flanger effect is applied to the alert signal, communicating when a high priority scenario is becoming more serious.

To accompany high priority alerts, a medium priority alert system relays when a robot’s hazard priority level passes the 0.5 threshold. These short earcons, based on the same sounds as high priority alerts, differ depending on whether the priority level is rising or falling.

\subsubsection{UI Feedback}
The final sounds emitted by robots are earcons resembling tonal `grunts’, with the tonal melody dependent on the situation. These are used to acknowledge user interactions. They also occur before a high priority alert begins sounding, creating a ``two-stage signal'', useful for conveying that a ``complex piece of information is about to be delivered'' \citep{case2018designing}. Feedback sounds for tagging, setting and removing waypoints, and confirming user interactions are stereo headlocked rather than spatialised.

\subsubsection{Ambience}
This element represents a first order ambisonic sound bed responding to the user's head movement. This aims to increase the sense of presence and helps reduce the detachment a user may feel when interacting with robots and objects within a cyber-physical model.

\subsubsection{Sonification Design}

The approach to sonification design proposed here follows the principles set out in section\ref{sec:CogMet}. One set of sounds (cog) utilise cognitive metaphor and ecological congruence in their design. 
A matching set of sonifications (comp) follow the archetypal, computationalist approach to sonification design.  Both sets of sonifications utilise a parameterised approach, whereby the value of a parameter is set according to the data level of the associated hazard. Fine details of the designs utilised are described in the following sections.We designed a comparitive study to evaluate the two, contrasting approaches. 

The \textbf{Cognitive Metaphor} (cog)  approach seeks 
to metaphorically express the physical dimensions of data sources, as well as being ecologically congruous with the conditions in which our auditory scene exists. Furthermore, it is important that when multiple sonification streams occur concurrently, each stream of the resulting auditory scene can be differentiated easily by a listener \citep{Neuhoff2002}. With these considerations in mind, cognitive metaphor and gestalt principles played a key role in our sound design and implementation. Each sonification aims to:
\begin{itemize}[itemindent=0.5cm, labelsep= 0.3cm]
    \item Clearly communicate the value of its underlying data feature by ``performing a [...] simulation of underlying physical phenomena'' \citep{dubus2013systematic}.
    \item Cohere to existing sonic associations users are likely to possess about the hazard \citep{walker2005mappings, cooper2018effects}.
    \item Maintain a singular discernible audio stream that maintains ``temporal coherence'' \citep{shamma2011temporal}, e.g., a consistent timbre and pitch, with sounds in the stream occurring continuously or in rapid succession.
    \item Integrate with other sounds emanating from robots to provide a cohesive character.
    \item Maximise listener comfort and minimise annoyance and fatigue (particularly alarm fatigue \citep{sorkin1988}).
\end{itemize}

The RTL radiation sonification (example \href{https://on.soundcloud.com/DtePg}{\color{blue}\underline{here}}) emulates the sound produced by a Geiger counter, consisting of short, high frequency clicks. Transient density varies in a rhythmically complex, yet subtle, way. Higher radiation levels are reflected by faster patterns of clicks. At very high levels, these transient clicks are interspersed with short chirps, designed to draw user attention 
.

The RTL sonification for flammable gas was designed to share the sonic characteristics with the sound of air inhalation or gas flowing in a pipe, the metallic quality of the sound corresponding with materials used in robotics. As can be heard \href{https://on.soundcloud.com/3r4L1}{\color{blue}\underline{here}}, higher gas density decreases the duration between sample onset 
, which results in the perceived sound of ``more gas’’. The sound was created using phased, reverse cymbals as source samples. These were passed through a phaser to create more variation over time. The samples were also filtered to remove low frequency content.

With the potential for three concurrent RTL sonifications, it was important to ensure the sonic qualities of each sonification remain unique and consistent, so gestalt principles took precedent in the design of the RTL temperature sonification. As heard \href{https://on.soundcloud.com/Y4uF6}{\color{blue}\underline{here}}, it consists of a simple sine wave pattern at a fixed pitch of 220Hz. Higher temperature increases LFO speed and FM amount 
. The rhythmic simplicity, smoothness of timbre, and fixed low-mid frequency of the sine wave contrasts with both the transient granularity of the radiation and atonality of the gas sonifications and their higher spectral content, facilitating auditory scene perception.

Priority alerts for each hazard are designed using unique source samples, inspired by, rather than derived from, their RTL counterparts. This is to ensure they do not become confused with the RTL sonifications. Whilst there is correspondence - \href{https://on.soundcloud.com/xoW3H}{\color{blue}\underline{radiation}} consists of clicks, \href{https://on.soundcloud.com/PjpW9}{\color{blue}\underline{flammable gas}} of noise and \href{https://on.soundcloud.com/d9AC7}{\color{blue}\underline{temperature}} of sine tones - high priority alerts are more melodic in nature, consisting of an ascending arpeggiated sequence. As Case and Day posit, ``a melody would be difficult to miss'' reinforcing our ``attention to such alarms rather than detracting'' \citep{case2018designing}. However, they go on: ``overt melody-making could run an additional risk of extreme annoyance from overuse''. For this reason arpeggiated notes are short and do not adhere to an explicit musical scale.

The \textbf{Computationalist Approach} (comp) has been developed in order to facilitate a fair comparison with the first sound set (cog). Hence, it is important that the comp sonifications avoid the inadvertent use of the novel approaches and considerations laid out for cog, whilst still adhering to an established and respected sonification framework. Fundamental in the design of comp is Dubus and Bresin’s systematic review of mapping strategies \citep{dubus2013systematic}. We use this extensive review of 179 sonification publications as a framework for sound design due to the authors’ motivation to exploit and organise ``the knowledge accumulated in previous experimental studies to build a foundation for future sonification works'' \citep{dubus2013systematic}. 
We consequently use the most common or successful auditory mappings based on the physical dimensions present in our system. It should be noted that in contrast to cog the sounds are designed with little attention paid to the application domain or user comfort.

As is the case for cog, `Loudness’ and `Spatialisation’ are reserved for `Distance’ and `Location’ respectively, to accurately convey the 3D environment of the cyber-physical model. 
Hence, we excluded these parameter mappings for use in our sonification design. The physical dimensions of the hazards present in our system are energy (radiation), density (flammable gas) and temperature; we utilised these physical properties in establishing appropriate mappings.

After Loudness, the next most common mapping for energy is pitch. Radiation is the primary hazard level we want to communicate and ``pitch is by far the most used auditory dimension in sonification mappings’', ``known to be the most salient attribute in a musical sound’’ \citep{dubus2013systematic}. The RTL radiation sonification for comp consists of a pure sine tone with pitch mapped over two octaves, as heard \href{https://on.soundcloud.com/vMTFG}{\color{blue}\underline{here}}. Mapping to higher frequencies is avoided; as Kumar et al have shown, ``sounds with high unpleasantness have high spectral frequencies and low temporal modulation frequencies'' \citep{kumar2012features}.

The RTL sonification for flammable gas, heard \href{https://on.soundcloud.com/VyraA}{\color{blue}\underline{here}}, utilises a white noise oscillator in accordance with Dubus and Bresin’s supported hypothesis that most sonification mappings follow the priorities of perception \citep{Dubus2013}. After pitch, they found duration to be the most common auditory mapping for density, and that the range between 100ms and 2s is used in more than 50\% of the reviewed studies. For this reason, the signal gain is attenuated using a sine LFO to create a beating effect, with frequency mapped between 0.5Hz and 10Hz. A sine LFO ensures the tone duration and inter-onset interval (IOI) are of equal length, no matter the LFO frequency, and that transitions between tone and IOI are gradual. This implementation method ensures the sonification is not mistaken as a notification at low gas levels.

The RTL sonification for temperature is mapped to signal brightness for two reasons. Firstly, brightness can be thought of as perceptually orthogonal to both pitch and beating, improving clarity between audio streams \citep{ziemer2020linearity} 
Secondly, after pitch and loudness – mappings reserved for radiation and distance – brightness is the next most successful auditory mapping for temperature in Dubus and Bresin’s sonification study review, being labelled to indicate that its ``efficiency was found to be significantly better when tested in comparison to other mappings corresponding to ``temperature sonification (2013). Our temperature to auditory brightness mapping is achieved using two sawtooth waves with fixed base frequencies of 55Hz and 110Hz, reducing spectral interference with the radiation audio stream. PWM with a duty cycle of 80\% is applied to the 110Hz signal to introduce further harmonic content in the upper frequencies. A 100Hz lowpass is applied at minimum temperature to filter out the higher signal and all harmonic content. The lowpass frequency is mapped to temperature level, reaching 20kHz at highest temperature. Finally, the gain of each oscillator is also mapped to temperature level, with the higher frequency signal being more prominent at high temperatures and vice versa. To the listener, this results in a consistent volume, with the sound being duller when the environment is cooler and brighter when hotter. It can be heard \href{https://on.soundcloud.com/mkf6D}{\color{blue}\underline{here}}.

As in cog, the comp \href{https://on.soundcloud.com/6DSCW}{\color{blue}\underline{notifications}}, \href{https://soundcloud.com/virtuallythere/sets/comp-high-priority-alerts/s-jq1DyaFes8G} {\color{blue}\underline{high priority}} and \href{https://on.soundcloud.com/pVzJZ}{\color{blue}\underline{medium priority alerts}} for each hazard type are modelled on their respective RTL sonifications, with UI feedback SFX and ambience remaining unaltered.

\section{User Study}
\label{sec:StudyDesign}
To evaluate our sonification approaches, a between subjects user study was conducted involving 50 participants (33 male, 16 female, 1 other, age = $M28.5 \pm 5.26$) randomly split into two equally sized experimental groups, with sound set as the independent variable (cog and comp). The study used the HTC Vive Pro Eye head-mounted display, with the Vive controllers for robot control and navigation, and using the built-in headphones for audio. The software ran on a PC with an Intel Core i9-7900X CPU @ 3.30GHz, with 32GB of RAM, and an NVIDIA Titan V graphics card.

Participants were tasked with identifying and labelling objects in the VR environment that they deem to be highly hazardous. To do this they must use RTL to determine hazard levels for objects within the environment and tag objects that are highly hazardous with the correct hazard type. Participants will need to redirect the robots to gather complete sensor data of some areas of the environment as the robot navigation route does not completely cover the area and hazard avoidance behaviour may also contribute to an incomplete data capture. Additionally participants are asked to keep the robots as safe as possible. This will involve setting waypoints around areas they perceive to be hazardous on the robot's exploration route, and minimising time inside hazardous areas that need additional sensor data capture. The scenario is kept constant across all participants.

\subsection{Study Protocol}
The study was undertaken at Bristol Robotics Laboratory based at UWE Frenchay campus, Bristol UK, and was divided into the following stages:
\begin{enumerate}[itemindent=0.5cm, labelsep= 0.1cm]
    \item \textbf{Instruction and consent:} Introducing the study purpose and itinerary and obtaining consent 
    \item \textbf{On-boarding:} Tasks, system operation and assigned sonification set were introduced using a video and 10 short tutorials in VR
    \item \textbf{Main Test Scenario:} Participants completed the tasks using the assigned sound set for a maximum duration of 20 minutes
    \item \textbf{Survey:} Gathering quantitative data about participants' experiences of the Main Test Scenario
\end{enumerate}

While the system is designed to be as self-explanatory as possible, there are a number of features that need to be understood for system operation. To streamline the learning of these features we designed a series of on-boarding (tutorial) scenes, conducted in VR, to enable participants to learn about system operation and data presentation. Each scene introduces a subset of system features, and allows the user to experiment with the system features presented within that tutorial.

\subsection{Dependent Measures}
In order to test our hypotheses on the efficacy of different approaches to data sonification we have used a variety of subjective dependent measures. In order to evaluate workload, trust in the system, and presence in the virtual environment we have used questionnaires that are common in the literature for assessing these facets of system performance:  the NASA task load index (NTLX) \citep{NTLX}, trust in automation (TinA) \citep{TinA}, and the iGroup presence questionnaire (IPQ) \citep{IPQ} respectively. In order to evaluate experiential factors directly related to the sonifications and their utility we have designed a set of questions presented in Table~\ref{tab:SEQs}. Finally we gathered qualitative feedback on the system as a whole to hopefully illuminate our findings from qualitative analysis.

\begin{table}
    \centering
    \footnotesize
    \begin{tabular}{|p{0.7cm}|p{6.5cm}|}
    \hline         
    SEQ1     & I found it easy to identify the gas sounds.\\ \hline
    SEQ2     & I found it easy to identify the radiation sounds.\\ \hline
    SEQ3     & I found it easy to identify the temperature sounds.\\ \hline
    SEQ4     & I was able to determine different hazard levels.\\ \hline
    SEQ5     & I found the alarms annoying.\\ \hline
    SEQ6     & I found the real-time listening sounds annoying.\\ \hline
    SEQ7     & I would be happy to have these sounds present all day if this were my job.\\ \hline
    SEQ8     & After a while I ignored the alarms.\\ \hline
    SEQ9     & After a while the sounds helped me understand the scene without needing to concentrate on them.\\
    \hline
    \end{tabular}
    \caption{Sound evaluation questions (SEQ). Each question is scored on a 5-point Likert Scale}
    \label{tab:SEQs}
\end{table}

Initially we had intended to log task performance metrics such as tagging accuracy and task performance time, however, during study pilots it immediately became clear that individual differences would result in noisy and useless data. Firstly, task interpretation was highly variable with some participants tagging everything slightly hazardous, some only the object at the centre of the hazard, others doing as requested and tagging highly hazardous objects. Despite improving the phrasing for the main study participants still interpreted the instructions quite differently. Secondly, there were large differentials in system operation capability which resulted in tagging errors that were not the result of sonification utility. Consequently, we conducted a secondary study to objectively evaluate the performance of the sonifications for data understanding (see Section~\ref{sec:2ndStudy}), while the main study allows us to evaluate subjective user experience of the sonifications in the VR environment they were designed for.

\subsection{Results}

Where data were normally distributed, independent t-tests were used for the comparison (t and p values are reported); however, in the majority of cases data were not normally distributed (Shapiro-Wilk tests with $p < 0.05$) so Mann-Whitney U tests were used (U and p values are reported).

For the NASA TLX data the mean (non-weighted) workload value was calculated for all contributory factors and compared using an independent measures t-test, no significant difference was found between the conditions $t=0.557, p=0.580$. Decomposing the NTLX into separate factors, mental workload was found to be significantly higher for cog ($M68.75 \pm 18.80$) than comp ($M53.85 \pm 19.61$), $U=189.5, p=0.015$. No other factors were found to be significant.

For the IPQ data the data was processed according to \cite{IPQ}, and thus analysed as presence (P), spatial presence (SP), involvement (I), experienced realism (ER). No factors of the IPQ were found to be significant.

For the Trust in Automation (TinA) data the data was processed according to \cite{TinA}, and thus analysed as reliability/competence (RC), understanding/predictability (UP), familiarity (F), intent of developers (IofD), propensity to trust (PtoT), and trust in system (T). UP was found to be significantly greater for comp ($M3.97 \pm 0.82$) than cog ($M3.65 \pm 0.64$), $U=418, p=0.038$. No other factors were found to be significant.

For the sound evaluation data each question was compared individually. The gas sounds were found to be easier to identify in comp ($M4.54 \pm 0.71$) than in cog ($M3.13 \pm 1.33$), $U=502.5, p=9.1e-5$. The radiation sounds were found to be easier to identify in cog ($M4.66 \pm 0.70$) than in comp ($M3.88 \pm 1.07$), $U=156.5, p=0.0009$. The alarms were found to be more annoying in cog ($M3.83 \pm 1.09$) than in comp ($M2.763 \pm 1.28$), $U=163.5, p=0.003$. No other results were found to be significant.

At the conclusion of the survey participants were invited to add “any additional comments”. Responses to this
question provide additional insights, particularly in relation to elements that users found challenging. Of the 50 study participants, 30 provided a free text response to this question, 15 of whom had experienced the cog condition and 15 who had experienced the comp condition. Twenty participants neglected to give a response to this question.

Although a relatively small sample set, there do appear to be some pronounced differences in the self-report of those experiencing cog and comp conditions. For our analysis we primarily used Linguistic Inquiry and Word Count (LIWC).

cog - User feedback is more focused on technical aspects, relating to areas that users felt could be practically improved (LIWC - tech cog: 1.93, comp: 0.2) The language used also suggests that greater attention was paid to the spatial qualities for users of cog (LIWC – space cog: 7.49, comp: 4.01) possibly inferring higher levels of situational awareness. cog participants also use more emotional language, both positive and negative in the way in which they report their experience (LIWC – emotion cog: 1.93, comp 1.0). Language such as ``overwhelmed'', ``disorientating'' and ``confusing'' is commonly used in negative feedback, ``fantastic'' and ``instinctive'' are used in positive feedback. 

comp - Feedback from those who experienced comp give multiple reports of sounds being difficult to tell apart from one another. Six of 15 participants who responded to this question having experienced comp reported difficulty in distinguishing between some combination of the radiation, temperature and flammable gas sonifications. None of the 15 respondents to this question who experienced cog expressed such concerns, however two recounted difficulties distinguishing between the more granular expressions of each sound e.g. high, medium and low levels of hazard. Negative language most commonly used to describe comp includes ``distracting'', ``mixed up'' and ``difficult''. Positive language included ``good'' and ``entertaining''.

\section{Sonification Evaluation Study}
\label{sec:2ndStudy}
\begin{figure}
    \centering
    \includegraphics[width=\columnwidth]{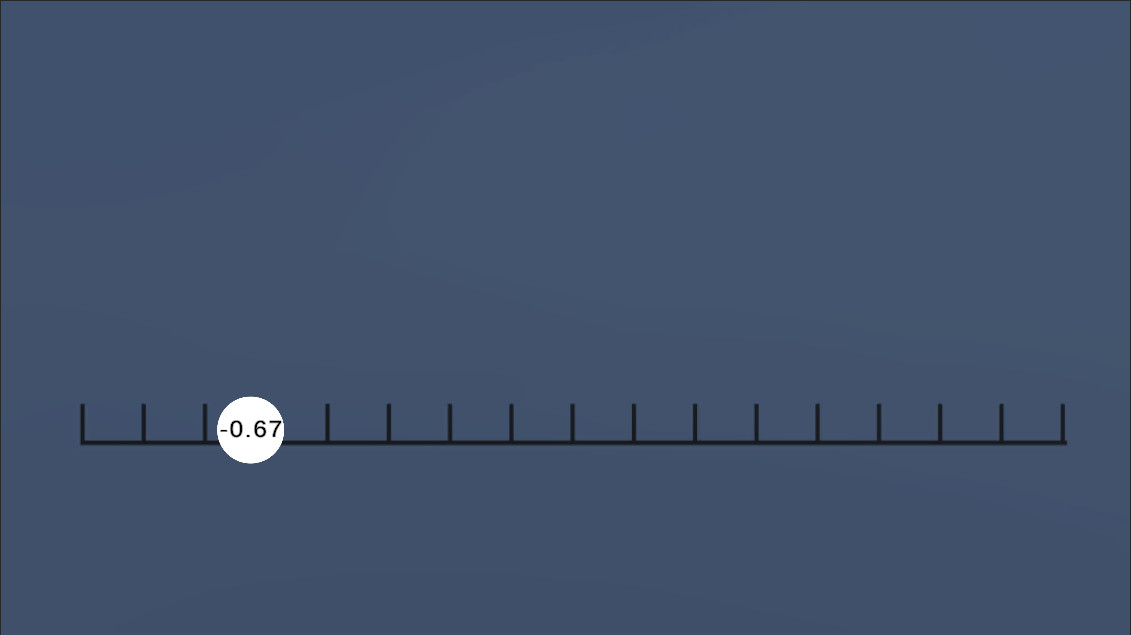}
    \caption{The video interface used for the sound evaluation study. The circle traverses the horizontal scale displaying its position.}
    \label{fig:SEVideo}
\end{figure}

To evaluate the efficacy of the sonifications for data representation, we have conducted an online study where participants had to identify the underlying data from its sonification. We used a within subjects design conducted on Prolific Academic, 50 participants were recruited (31 female, 17 male, 2 removed for giving random answers, age = $M43.85 \pm 14.32$), and compensated £4.50 for their time, recruitment was restricted to those without a hearing impairment, and they were instructed to wear headphones for the study. For each sonification participants were tasked with observing a video in which data values changed according to a hidden function, and had to identify at which point the sonifications indicated a maximum level for the data. For this task a set of videos were created in which a circular marker moved across the frame from left to right and back again, while displaying the numerical value of its position (Figure~\ref{fig:SEVideo}), a hidden Gaussian function converts that location into a data value that was then sonified using the RTL approach described in section~\ref{sec:audio} (there was no spatialisation so the audio was equivalent to self-RTL); the centre for the Gaussian in each video was randomly generated in the range -1 to +1. Each question block relates to one of the 6 sonifications, and consists of a set of 3 videos with randomised maximum values, displayed in a random order. To answer each question participants were permitted to re-watch the video as required, and then had to pause it at the location for which the data was at a maximum and enter the position value displayed. At the end of each block participants had to identify which hazard they believed the sound represented.

\subsection{Results}
\begin{figure}
    \centering
    \includegraphics[width=\columnwidth]{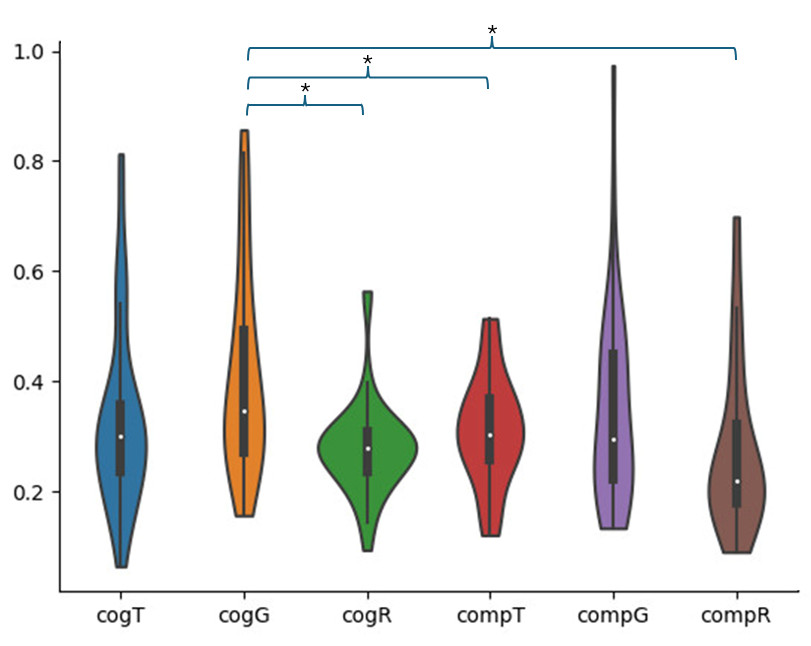}
    \caption{Results distributions for the mean errors in data-level identification. Gas (G), Temperature (T) and Radiation (R). Significant comparisons are indicated with a *}
    \label{fig:SE_Results}
\end{figure}

To analyse the data for the hazard level identification accuracy, participant responses for each question were subtracted from the hazard maximum, and the absolute values for these calculated distances were then averaged for the three trials for each sound for each participant. On observation of the results distribution it was apparent that there were a number of outliers that would impact subsequent statistical analysis. Hence, outliers more than 3 standard deviations from the mean were filtered out prior to statistical analysis, this reduced the participant sample to 40 participants. To analyse the data we used a 2-way repeated measures ANOVA. However, the data violates the normality assumption (Shapiro-Wilk tests show that for 5 out of 6 data groups $p<0.05$), with excessive skew ($ABS(skew) > 2$ for some conditions). To compensate for the excessive skew the data is transformed using a square root transform. The results following this data processing are shown in Figure~\ref{fig:SE_Results}. For higher order ANOVA ranked approaches have challenges for post-hoc comparisons, and other non-parametric alternatives (e.g., ART-ANOVA) are not fully investigated. As such we have relied on the robustness of ANOVA to normality violations, and the fact that our transformed data is within reasonable bounds for skew ($-2 < skew < 2$ for all conditions) and kurtosis ($kurtosis < 5$ for all conditions); hence, our analysis and statistical conclusions are still valid.

Maulchy's test indicates the data adheres to the assumption of sphericity $\chi^2(2)=1.812, p = 0.404$. A 2-way repeated measures ANOVA shows that there is a main effect of hazard type ($F(0.41, 2)=8.23, p=0.0006$), but no main effect of sound-set ($F(0.06, 1)=2.66, p=0.11$) or interaction effect ($F(0.04, 2)=1.16, p=0.31$). As there is no effect of sound set or interaction effect, pairwise comparisons are conducted between all 6 sounds (grouping the data by hazard type and ignoring sound set would not facilitate useful evaluation). Due to the large number of comparisons we have used False-Discovery Rate (FDR) with $alpha=0.05$ to compensate for the multiple comparisons \citep{storey2011false}. The results of the analysis (Figure~\ref{fig:SE_Results}) show that the gas sonification from cog ($M0.41 \pm 0.18$) performs significantly worse than cog radiation ($M0.18 \pm 0.09$), $t(39)=4.30, p=0.0016$; comp temperature ($M0.31 \pm 0.10$), $t(39)=3.457, p=0.0066$; comp radiation ($M0.27 \pm 0.15$), $t(39)=3.49, p=0.0066$. All other comparisons were not significant.

\begin{figure}
    \centering
    \includegraphics[width=\columnwidth]{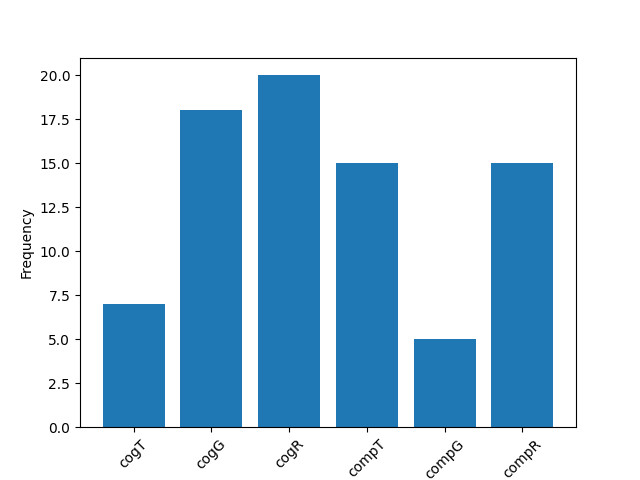}
    \caption{Frequency with which the data types were correctly identified. Gas (G), Temperature (T) and Radiation (R).}
    \label{fig:soundID}
\end{figure}

\begin{figure}
    \centering
    \includegraphics[width=\columnwidth]{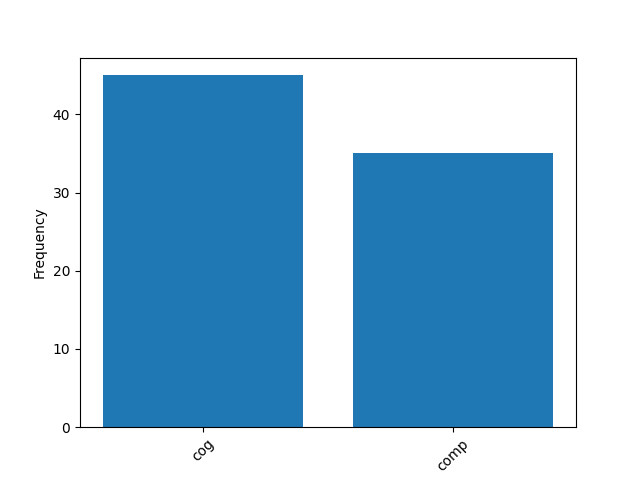}
    \caption{Frequency with which the data types were correctly identified grouped by design approach.}
    \label{fig:soundIDSS}
\end{figure}

In order to analyse the hazard identification data, we have used used a Cochran's Q omnibus test to investigate if there are differences in the number of hazards identified correctly for the different sound designs. To do so the data has been binomially coded (i.e., as wrong or right). The test shows that there is a significant difference between sounds $\chi^2(5)=19.42, p = 0.002$, hence, we conducted pairwise Cochran's Q-Tests (with FDR correction, $alpha=0.05$) between all of the sounds. We found that sounds cogT and compG were harder to identify than the other sounds. The results are shown in Figure~\ref{fig:soundID}. Grouping the sounds by design approach we found there was no significant difference in hazard identification between the approachs. The results are shown in Figure~\ref{fig:soundIDSS}.

\section{Discussion}

The central finding from the results suggest that when comparing cognitive metaphor (cog) and computationalist (comp) sound design approaches there was little impact on subjective experience or utility for data understanding. Where differences were found the results tended to favour the more conventional approach to sonification design. However, particularly in light of this being contrary to our expectations, it is instructive to analyse our findings, how they relate to our hypotheses, and by reasoning on them suggest avenues for further research. 

H1 suggests that cog sounds would give users an increased sense of presence, and was refuted. We posit that engaging the auditory channel regardless of sound design was sufficient for an increased sense of presence, as all factors of the IPQ were positively skewed for both sound design approaches. This finding corroborates \citep{larsson2007you}, and it seems reasonable to suggest, that while our design intention was for the sounds to be better situated in the environment, they could not be made sufficiently more congruent, while maintaining data representation, to have a positive effect on presence. Indeed, they may not be sufficiently aligned with the participants associations and expectations for the hazards types \citep{Kantan2021}. However, this finding does suggest that using either sonification approach would lead to task performance increases connected to an increased sense of presence \citep{bowman2007virtual}, while not resulting in complete immersion that may not be desirable in a hazardous environment \citep{malbos2012behavioral}.

The absence of any differences in measures relating to trust between the design approaches, refuting H2, are directly linked with the absence of differences in presence \citep{Salanitri2016}. Further, as the scenario was entirely simulated, beyond risks to the participants' task performance (of keeping the robots safe), participants may have seen it as a low risk scenario reducing their need to trust in the system. In future work we intend to have a real robot fleet which mirrors the actions of the simulated robots, and investigate what implications there might be for trust in a system with real-world consequences.

As with H1 and H2, H3 was similarly refuted with a lack of evidence that designing sounds which allow the application of acquired contextual knowledge are easier to interpret and utilise. Indeed, the opposite was found for the mental workload component of the NTLX. We think this opposing finding is linked to the reduced predictability of the cog sounds relative to the comp sounds, as participants had to pay more attention to the cog sounds as they were less sure about how they would change. However, the findings from the qualitative data suggest that the cog sounds were more intuitive and had better spacialisation so could be localised more easily. To reconcile these conflicting pieces of evidence, the qualitative data suggests that the novel sonic landscape, and complex system usage may have obfuscated any difference in the impact on cognitive load of the relative intuitiveness of the different sound design approaches. Consequently, for future studies we intend to evaluate the systems over successive interactions to allow users to become naturalised to the sounds and system operation. The change over time of task performance, and final cognitive load of the system will be instructive as to the utility of different design approaches.

\begin{figure}
    \centering
    \includegraphics[width=\columnwidth]{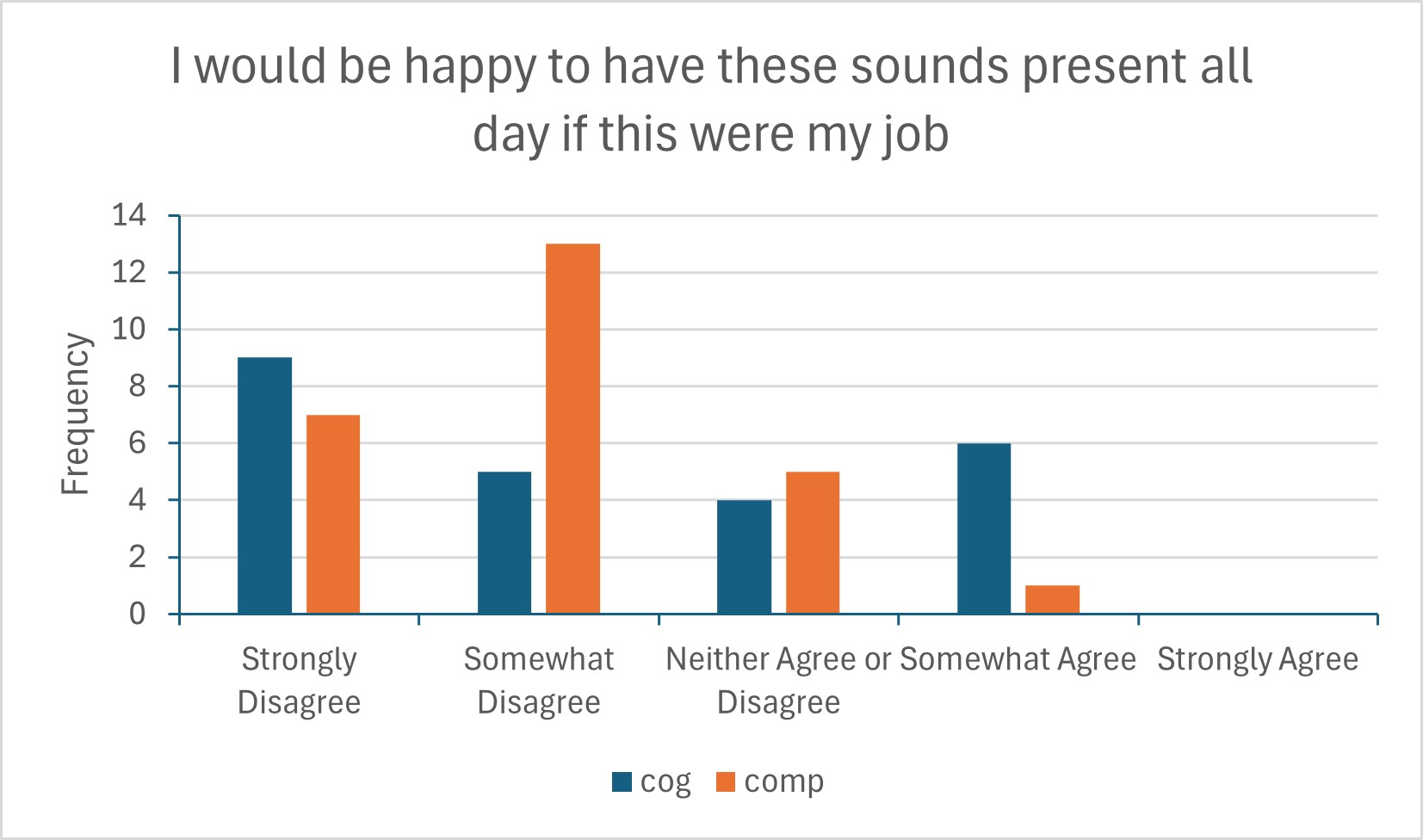}
    \caption{Results of SEQ7.}
    \label{fig:SEQ7}
\end{figure}

We found that there was no difference in subjective listening experience between the two RTL sound-sets, and the notification sounds for the cog sounds were found to be more annoying. These findings refute H4. However, looking at a graph of the results for SEQ7 (Figure~\ref{fig:SEQ7}), there is a clear trend for participants to prefer using the cog sounds as part of their job: 77\% disagreed they would be happy to listen to the comp sounds all day and only 4\% agreed, whereas 58\% disagreed for the comp sounds and 25\% agreed. Two factors suggest this observation might be of use: firstly with more participants this difference may become significant, secondly, and more importantly, in future studies with longer exposure times to the sounds these results are likely to shift, becoming more pronounced. It is also noteworthy that modifications may need to be made to sonification design for users to be happy with workplace usage of data sonification, though this is caveated with the fact that users were exposed to the sounds for only a short time period.

In the follow-up study we observed that how easy it was to connect sounds to data type was dependent on individual sound design choices rather than the design approach, refuting H5. By following the suggestions in the literature for physical property to sound mappings the hazard types were as easy to identify as those designed with cognitive metaphor. It is apparent that participants were as able to connect the mappings of sound parameters to physical properties as suggested in \citep{Dubus2013} for the Comp sounds, as they were able to associate the contextual appropriateness and utilise their embodied knowledge \citep{Roddy2020} for the cog sounds.

The results of the follow-up study also demonstrated that due to adhering too closely to cognitive metaphor, data levels of the gas sound from that sound-set were harder to determine than for any other sound, partially confirming H6. However, the qualitative data from the main study demonstrated that participants experiencing the comp sounds found it harder to discern data levels when multiple sounds played simultaneously. This provides some validation to the cog approach where sounds are carefully designed to be used simultaneously. This finding suggests that should the comp design approach be more appropriate for a particular setting, the sounds could be improved by designing the sounds to better suit simultaneous play. To do this the sounds may well need to deviate from the proposed mappings, i.e., a blend of the two approaches to sonification design. 

\section{Conclusion}
In this paper we have detailed a novel approach to data sonification based on cognitive metaphor and ecological congruency (cog), applied to a virtual reality robot teleoperation system for use in nuclear decommissioning. While prior work has applied those principles to sound design generally, to the best of our knowledge prior work has not applied them to sonification design. Further, we evaluate this approach through experimental comparisons with a computationalist approach to sonification design (comp) based on recommended sound mappings in the literature. Our main finding was that the computationalist approach performed slightly better than the cognitive metaphor approach on a small subset of metrics, principally on predictability and mental workload. However, qualitative data analysis demonstrated that the cog approach resulted in sounds that were more intuitive, and were better implemented for spatialisation of data sources and data legibility when there was more than one sound source. 

Our results also highlighted the need for more prolonged testing periods so that users could become more naturalised to system operation and the soundscape. Such longer form studies would allow formation of a better picture of the impact of design decisions on task performance and subjective experience without the obfuscation of system complexity. Further, the impact on system learning time and prolonged exposure effects on noise fatigue would both be valuable factors to study. It is important to note that a limitation of our findings is that the objective evaluation of the sonifications was done outside of the context of VR so our conclusions on their effectiveness in this context less definitive. Our proposed extended user testing protocols would allow us to effectively evaluate them in a VR context without the obfuscation of individual user differences.

\section{Acknowledgements}
We would like to thank Prof. Paul White for providing help with the statistical analysis and Philip Tew for supporting the development of the virtual environment. This research has been made possible thanks to funding from the UKRI Trustworthy Autonomous Systems Hub (EP/V00784X/1).

\bibliography{jaes.bib}
\bibliographystyle{Frontiers-Harvard.bst}

\end{document}